\documentclass{article}
\usepackage{ijcai26}

\usepackage{times}
\usepackage{soul}
\usepackage{url}
\usepackage[hidelinks]{hyperref}
\usepackage[utf8]{inputenc}
\usepackage[small]{caption}
\usepackage{graphicx}
\usepackage{amsmath}
\usepackage{amsthm}
\usepackage{booktabs}
\usepackage{algorithm}
\usepackage{algorithmic}
\usepackage[switch]{lineno}

\title{GRACE: Graph Neural Networks for Locus-of-Care Prediction\\ under Extreme Class Imbalance}

\author{
Subham Kumar\textsuperscript{\dag}
\and
Lekhansh Shukla\textsuperscript{\ddag}\and
Animesh Mukherjee\textsuperscript{\dag}\and
Koustav Rudra\textsuperscript{\dag}\And
Prakrithi Shivaprakash\textsuperscript{\ddag}\\
\affiliations
\textsuperscript{\dag}Indian Institute of Technology, Kharagpur\\
\textsuperscript{\ddag}National institute of mental health and Neuro Sciences, Bangalore\\
\emails
\{kumarshubham209, drlekhansh, animeshm, krudra5, prakrithishivaprakash\}@gmail.com,
}

\usepackage{graphicx}
\usepackage{multirow}
\usepackage{tabularx}
\usepackage{subfig}
\usepackage{lineno}
 \usepackage{amsmath}
\usepackage{booktabs}
\usepackage{paralist}
\usepackage{arydshln}

\usepackage[table]{xcolor}
\definecolor{lightgreen}{HTML}{90EE90}
\usepackage{soul}

\usepackage{colortbl}
\newcommand{\subham}[1]{\textcolor{blue}{[SK: #1]}}

\newcommand{\ta}[1]{\texttt{T\textsubscript{A}}}
\newcommand{\tb}[1]{\texttt{T\textsubscript{B}}}
\newcommand{\tc}[1]{\texttt{T\textsubscript{C}}}
\newcommand{\ours}[1]{\textsc{Grace}}
\newcommand{\grgcn}[1]{\textsc{Grace\textsubscript{GCN}}}
\newcommand{\grgsg}[1]{\textsc{Grace\textsubscript{Gsage}}}
\newcommand{\grgat}[1]{\textsc{Grace\textsubscript{GAT}}}
\newcommand{\grgt}[1]{\textsc{Grace\textsubscript{Gtran}}}
\newcommand{\grtron}[1]{\textsc{GraceTron}}
\newcommand{\fb}[1]{$\mathcal{F}_B$}
\newcommand{\fe}[1]{$\mathcal{F}_E$}
\newcommand{\fl}[1]{$\mathcal{F}_L$}
\newcommand{\fr}[1]{$\mathcal{F}_R$}

\begin{document}
\maketitle
\begin{abstract}
Determining the appropriate locus of care for addiction patients is one of the most critical clinical decisions that affects patient treatment outcomes and effective use of resources. With a lack of sufficient specialized treatment resources, such as inpatient beds or staff, there is an unmet need to develop an automated framework for the same. Current decision-making approaches suffer from severe class imbalances in addiction datasets. To address this limitation, we propose a novel graph neural network (\ours{}) framework that formalizes locus of care prediction as a structured learning problem. In addition, we propose a new approach of obtaining an unbiased meta-graph to train a GNN to overcome the class imbalance problem. Experimental results with real-world data show an improvement of 11-35\% in terms of the F1 score of the minority class over competitive baselines. Further, if we jointly finetune the base embedding fed into \ours{} as input together with the rest of the GNN component of \ours{}, there is a remarkable boost of 15.8\% in performance.
\end{abstract}
\section{Introduction}
Over the past decade, several strategies for clinical decision support system~\cite{cdss} have been introduced to enhance patient care management.
To contextualize the objectives of this work, it is important to understand the domain of addiction psychiatry and the unsolved challenges. One such challenge is in determining the appropriate decision for patient care based on clinical condition. Substance use disorders (SUD), especially alcohol use disorders (AUD), cause brain damage and premature death \cite{Volkow2023}. For a person with AUD to stop drinking is often fraught with danger, characterized by psychosis and complicated withdrawal symptoms (seizures, delirium, etc.). Half of those who suddenly stop or reduce their drinking tend to experience alcohol withdrawal syndrome, although the severity varies~\cite{Goodson2014AWS}. 
The significance of this problem was acutely realized during COVID-19, when a large number of patients developed complicated withdrawal syndrome in India \cite{covid19}. In addition, \cite{unknown} showed 90\% unmet need for treatment of SUDs in low-resource countries. In the real world, this presents as a problem of triage and task-sharing. The process of safely stopping alcohol use requires medical treatment, which is commonly called as detoxification. Healthcare providers must classify patients seeking alcohol detoxification into \textit{inpatient} (IP) and \textit{outpatient} (OP) triage.
In medical systems, this decision point is referred to as the `locus of care'. 

\begin{figure}[t]
  \includegraphics[width=\linewidth]{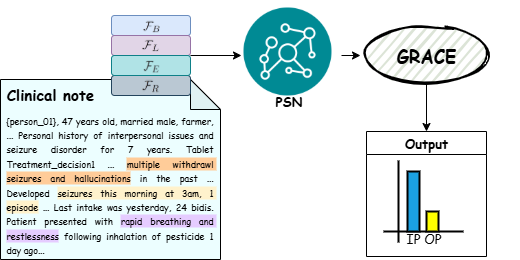}
  \caption{\footnotesize An example illustrating \ours{} framework as a black box, given an input clinical note with the reasoning pathways highlighted.}
  \label{fig:clinical_example}
\end{figure}

Computationally, this task of binary classification is challenging due to the highly imbalanced nature of the addiction dataset. In the real world, inpatient cases are significantly lower than outpatients, primarily due to two reasons. First, most AUD patients actually do not develop severe illness. Second, there are resource constraints in terms of the number of experts to effectively manage patients or the availability of hospital infrastructure to monitor admitted patients. 
A meta analysis on alcohol withdrawl syndrome~\cite{Goodson2014} investigates predictors such as platelet count, blood alcohol levels, and liver function tests. Laboratory tests are often unavailable in low-resource centres, and when available, have a turn-around time of 3-4 hours at least, precluding efficient and quick decision-making in a busy outpatient setup. However, information collected during a clinical consultation itself (mostly recorded digitally as a \textit{clinical note} by medical experts) can provide signals to assess the risk of complicated withdrawal. \\
This work is motivated by the growing need to develop an accurate and practically applicable clinical decision support system in addiction treatment to predict the locus of care. To address these gaps~\cite{gaps}, in this paper, we propose a GNN-based unified framework -- \ours{}. The novelty lies in (a) obtaining high dimensional numerical representation of the patient (section \ref{subsec:node_features}), (b) expressing the `semantic' similarity between patients based on these features, (c) feeding this network to a meta-learning anchored GNN model to make accurate predictions (section \ref{subsec:metaGNN}). In addition to the traditional textual features extracted from the clinical note, we also featurise the reasoning pathways that the clinician potentially uses to arrive at the present locus of care decision. Typically, when a clinical note is drafted about a patient, the medical expert follows a reasoning pathway to arrive at the present locus of care decision. Some of this reasoning might be explicit in the notes, while others might be implicit. We make use of the SOTA reasoning-based large language models (LLMs) to extract the reasoning `hidden' in the clinical notes (see Figure~\ref{fig:clinical_example} for an example). We fuse this silver data as an additional node feature with the objective of improving the predictive power of our model.
As a final innovation, we introduce a novel variant of \ours{} where we jointly finetune \textsc{GatorTron}~\cite{gatortron} language model embeddings (LME) along with GNN using the same loss function backpropagating errors to both GNN and LME, enhancing the predictive power of the \ours{}. The idea is to obtain embeddings not only from finetuning LME but also `teach' it the graphical properties of patient nodes. \\
\noindent\textbf{\textit{Contributions.}}
This paper makes the following contributions.
\begin{compactenum}
    \item To our knowledge, \ours{} is the first model to predict locus of care decisions for addiction patients.
    \item We conduct extensive evaluations on \ours{} using different GNN architectures, including \textsc{GCN} \cite{gcn} (\grgcn{}), \textsc{GraphSAGE} \cite{graphsage} (\grgsg{}), \textsc{GAT} \cite{gat} (\grgat{}), and \textsc{GraphTransformer} \cite{graphtranformer} (\grgt{}).  \grgsg{} demonstrates the optimal performance, in terms of F1 score (0.74 for the minority class), followed by \grgt{} (0.73) (see Table~\ref{tab:ipop}) when compared with latest works ~\cite{deepj,graphgpt,patemb} and SOTA LLMs in zero-shot setting by 11-35\%. 
    \item Next we propose a setup to jointly finetune the \textsc{GatorTron} embeddings~\cite{gatortron} that are (already) input features to \ours{} along with the rest of the GNN framework of \ours{}. The final loss is backpropagated through the GNN as well as the top layers of \textsc{GatorTron} so that the initial embeddings have a scope to adjust to the task at hand. This variant of \ours{} achieves the overall best performance of 0.95 in terms F1 score of the minority (IP) class. This beats SOTA medical LLMs by 15-26\%.
    \item We perform ablation of node features and meta-graph (section~\ref{sec:results}). Sequentially enriching node features led to an absolute net gain of 2-23\%, while omitting meta-graph reduced the performance by 7\%. 
\end{compactenum}
\noindent\textit{\textbf{Impact.}} Clinically, the deployment of our work would support the clinicians and healthcare providers for quick and reliable decision, thereby optimising resources and reducing adverse outcomes. This work will also set a foundation for novel, cost-effective AI-Assisted triage, enhancing healthcare delivery, especially in low- and middle-income countries with limited resources. The findings from this research work, along with the algorithmic architecture, not only offer a blueprint for developing AI-assisted decision support in addiction psychiatry but across various other domains. Finally, upon acceptance, we shall publicly release the enriched embeddings for all patients used in this study, which can be used in further or related studies.

\section{Related work}
\label{sec:related}
\noindent\textbf{Traditional machine learning}:
The emergence of machine learning (ML) presented a new beacon of hope during the early 2010s in clinical decision-making. Researchers in this decade used classical ML \cite{MLsurvey} algorithms like logistic regression ~\cite{sud2017use} 
random forests \cite{aud-dss2023}, and support vector machines (SVMs) \cite{SVM2015} as potential tools for better diagnosis of substance use disorders \cite{SUDML,aud2013increased}.\\ 
\noindent\textbf{Graph neural networks}: The introduction of GNNs \cite{GNNlit2020} resulted in a significant breakthrough in healthcare informatics. While traditional methods treat patients as independent data points, GNNs \cite{weightedPSN2021} model them as networks of interacting nodes. Patient similarity networks \cite{PSN2018} represent an emerging paradigm in precision medicine that utilises network structures to cluster patients on the basis of complex and heterogeneous features, such as genomic profiles and clinical attributes. Using methods such as similarity network fusion and netDx \cite{netDx} allows patient stratification in an interpretable, accurate, and reproducible manner. Although direct applications to the placement of addiction care remain limited, PreciseADR \cite{preciseADR} and LIGHTED \cite{lighted} show testing grounds for heterogeneous GNNs that meld multi-type nodes and temporal sequences to enhance adverse drug reaction or opioid misuse risk prediction accuracy. In the recent work ~\cite{deepj} used graph convolutional transformer with differential pooling for temporal modelling of patients. \\ 
\noindent\textbf{Large language models}: Concurrently, the rise of techniques empowered by LLMs \cite{llmSUD} such as BERT and GPT created new frontiers for classification in addiction care. LLMs illustrated exceptional capabilities at knowledge extraction from unstructured clinical notes, patient communications, and even social media \cite{llmSUDsocialmedia} stories on addiction and recovery. ~\cite{patemb} developed technique to obtain embeddings of diagnosis codes and progession notes for disease onset prediction. 
Recent research has accepted these advances and embraced the use of adaptive systems powered by reinforcement learning (e.g., Q-learning \cite{adaptivetreatmentQlearning}) to alter treatment intensity in real-time according to addiction patient response. 
\section{Dataset}
The data for this study comes from a tertiary teaching hospital with a specialised addiction treatment centre offering 24-hour emergency services, a thrice-weekly outpatient clinic, and an 80-bed ward. A team of clinicians and developers co-created an electronic health record (EHR) for addiction services, enabling outpatient services to become paperless as of January 1, 2018. The EHR of each visit includes structured fields for substance use (classes, quantity-frequency, last use, etc.) and free-text clinical notes, $\mathcal{N}$. There are a total of 147,230 entries in $\mathcal{N}$. Further, we denote $\mathcal{N}_p^i$ as the clinical note for $i^{th}$ visit of the patient $p$. In addition, we collate these $\mathcal{N}_p^i$ notes for a patient over the $i$ visits to obtain $\mathcal{N}_p$.
The medical records of all patients who sought treatment between January 1, 2018, and December 31, 2025, comprise the universe of this study. Patients are included in the current study if they fulfilled the following criteria.
\begin{compactenum}
    \item  Clinical diagnosis of mental and behavioural disorders due to use of alcohol, i.e., International Classification of Diseases version 10 codes F10 \cite{WHO1993ICD10}. 
    \item At least one visit where medical detoxification was prescribed (Lorazepam, Diazepam, Chlordiazepoxide) and/or at least one visit where the patient was admitted.
\end{compactenum}   
Following this selection, we have used only the timestamped entries from  $\mathcal{N}$ 
for training the model. This leads to a total of 55,587 entries in $\mathcal{N}$ from 9,296 patients. The basic question we attempt to answer is whether it is possible to automatically predict the type of care (IP vs OP) needed by a patient $p$ given $\mathcal{N}_p$. \\ 
\noindent\textbf{Cleaning and standardisation of the dataset}: These entries in $\mathcal{N}$ were entered by different doctors over a seven-year period and are riddled with non-standard abbreviations, agrammatism, and jargons. To correct all the entries in $\mathcal{N}$, we use the pipeline documented in  \cite{note_correction}, where the authors finetune a Llama-3~\shortcite{web4} model for clinical note correction.\\
\noindent\textbf{Removal of personal identifiable information}: While we did not use any sociodemographic variables from the EHR database for building our model, we realised that the entries in $\mathcal{N}$ themselves contain substantial personal identifiable information (PII). Using 1500 annotated examples we finetuned a BERT based NER model \cite{gliner} (character-level recall of 1.0 and precision of 0.98) and used it to identify and mask the PIIs across our whole dataset.
\if{0}
\begin{compactitem}
    \item Person (name without title or designation).
    \item Name of languages.
    \item Groups (tribal, religious, self-help, political).
    \item Company (names of healthcare facilities or places of employment).
    \item Dates (only fully specified dates or time periods).
    \item Numerical identifiers (hospital identification numbers or any other numerical identifiers that can be tied to a unique individual).
    \item Address (name of geographical entities, including country, state, city or locality).
\end{compactitem} 
 Out of these, 1,500 annotated entries were 
used to finetune a BERT based NER model \cite{gliner} for extracting the entities mentioned above. The performance of the finetuned model on a held-out set of 350 entries was found to be satisfactory, with character-level recall of 1.0 and precision of 0.98. There are a total of 8,513 entities detected and removed from these 1,850 entries (see Appendix~\ref{sec:dataset_analysis} for detailed statistics and performance of extraction of the different entities).\\ 
\fi
\noindent\textbf{Masking of target leaks}: The entries in $\mathcal{N}$ can contain information which directly reveals the outcome of the consultation, for example, ``Admit in Male Ward'' or ``home-based detox''. There can be multiple variations of these, and thus, we developed a systematic method to remove them. Addiction specialists annotated 3,250 entries which was used to train a specialised BERT model for indentification of leaked tokens. We use this model (macro-F1 0.88) to identify all occurrences of leaked tokens across the dataset and masked them.
\noindent\textbf{Final dataset}: At the end of the above process, we have 7,628 patient notes in $\mathcal{N}_p$. These notes are time-ordered and divided into train and test splits based on the recency of visits. The training set consists of patients who have completed all their visits before $1^{st}$ Jan 2023, and the test set has patients whose visits started on or after $1^{st}$ Jan 2023. This ensures that there is no scope for data leakage. With this split, we have 4,988 and 2,640 patients in the train and test splits, respectively. Given an instance of this dataset, the task attempts to predict the locus of care (binary IP vs OP classification) for the patient. \\

\section{Methodology}
\label{sec:method}
\begin{figure*}[t]
\centering
\small
  \includegraphics[width=\textwidth]{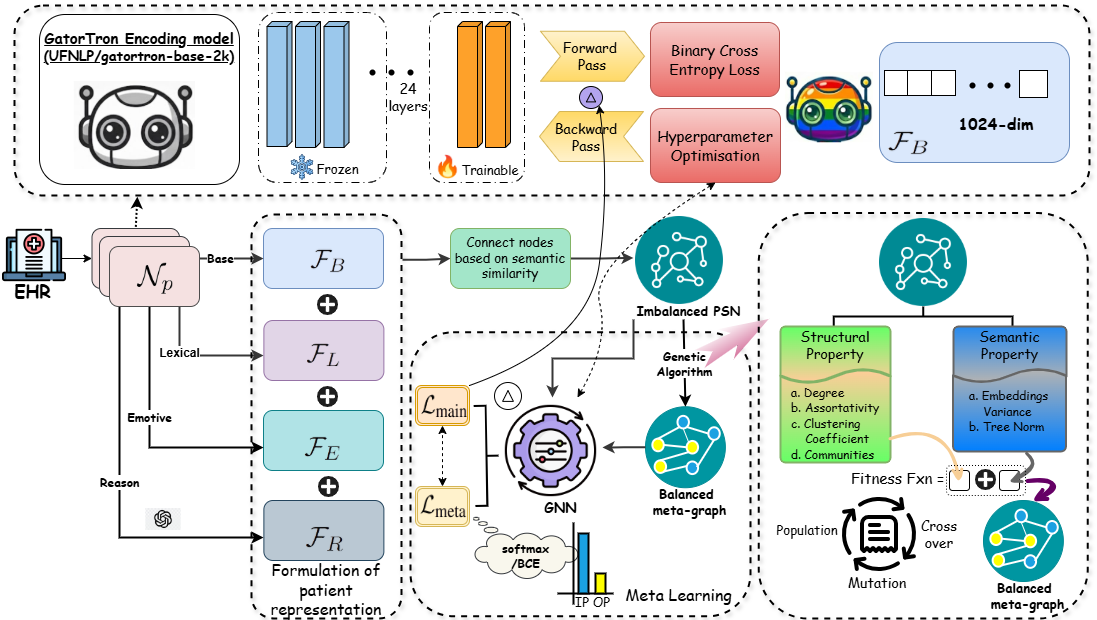}
  \caption { Workflow of \ours{} framework. Recall that $\mathcal{N}_p$ represents all the visit notes of patient $p$ concatenated together. \fb{}, \fl{}, \fe{}, and \fr{} represent the components of node features as base, lexical, emotive, and reason embeddings, respectively.}
  \label{fig:frmk}
\end{figure*}

This section describes the \ours{} framework to predict the locus of care for a given $\mathcal{N}_p$. This is a two-step framework including (i) formulation of node representation for each patient from $\mathcal{N}_p$ and (ii) the construction of the patient-similarity network, followed by the training of the meta-learning anchored GNN. The entire worflow is prsented in Figure~\ref{fig:frmk}. 

\subsection{Formulation of patient node representation}
\label{subsec:node_features}
Give a patient note $\mathcal{N}_p$, we featurize it by extracting multiple types of embeddings from it.\\
\noindent\textit{\textbf{Base embedding}}: We pass each $\mathcal{N}_p$ through a sentence transformer (or some other language model embedding (LME)) to obtain a 384-dimensional dense vector. This constitutes the base representation for an $\mathcal{N}_p$ corresponding to a patient $p$, and we call this feature as \fb{}.\\
\noindent\textbf{\textit{Lexical features}}: We enrich the base embeddings \fb{} by concatenating $n$-gram (lexical) features (\fl{}). The main goal of having these features is to find and use discriminative $n$-grams, specifically trigrams, that are statistically indicative of each class. For this, we compute the log-likelihood~\cite{llrtfrombook} comparing the goodness of fit of the data with two competing hypotheses \cite{lrt} mentioned below:
\begin{compactenum}    
    \item Null hypothesis ($\mathcal{H}_0$): The occurrence of the trigrams is independent of the patient class. In other words, the probability of observing the trigram is the same for both the IP and the OP classes.
    \item Alternative hypothesis ($\mathcal{H}_1$): The probabilities of the trigrams occurring differ between the IP and the OP classes.
\end{compactenum}
Only those trigrams are retained for which the $p$-value of the test is $<0.01$. From this exercise, we obtain a total of 723 trigram features, out of which 480 are distinctive of the IP and 243 of the OP classes, respectively. Thus the total embedding size is $|$\fb{}+\fl{}$| = 1107$.\\
\noindent\textbf{\textit{Emotive features}}: We use the \texttt{empath}~\cite{empath} library to extract emotive features (\fe{}) from each $\mathcal{N}_p$. The library has a broad set of pre-defined 194 emotional and topical categories, including \textit{anger}, \textit{confusion}, \textit{death}, \textit{fear}, \textit{injury}, \textit{sadness}, etc. Each category has a dictionary of words that correspond to the overall emotion/topic expressed by that category. As a result, the total embedding size for a patient node now is $|$\fb{}+\fl{}+\fe{}$| = 1301$.\\
\noindent\textbf{\textit{Reasoning pathways}}: We obtain the reasoning pathways given an input $\mathcal{N}_p$ by prompting a reasoning-based LLM. We encode these reasonings with the same sentence transformer as that for the base embedding. We then concatenate these dense 384-dimensional reasoning embeddings (\fr{}) with the node representation obtained so far and finally construct a feature vector of size $|$\fb{}+\fl{}+\fe{}+\fr{}$| = 1685$. The prompt to obtain a reasoning text from $\mathcal{N}_p$ is mentioned in \textit{supplementary materials}.
\if{0}\begin{figure}[!htpb]
    \centering
    \includegraphics[width=\linewidth]{Figures/PSN1.png}
    \caption{Formulation of Patient Similarity Network based on patient notes $\mathcal{N}_p$.}
    \label{fig:psn}
\end{figure}\fi


\subsection{Meta learning anchored GNN}
\label{subsec:metaGNN}
\noindent\textit{\textbf{Construction of the patient similarity network}} (PSN): Each patient node ($p_i$) in PSN is a 1685-dimensional vector, and the edge between two nodes $p_i$ and $p_j$ expresses the extent of similarity between the corresponding two patients. In particular, two patient nodes in PSN are connected if the cosine similarity between their vectors is $>0.8$.\\  
\noindent\textbf{\textit{The meta-learning framework}}: Recall that our dataset is mainly imbalanced. To address this imbalance in the dataset, we employ a meta-learning technique \cite{metaGCN} that adaptively modifies weights based on a small, balanced meta-graph. The meta-learning technique allows the model to change its learning process on the fly. The main idea of this method is that the model not only learns to classify nodes but learns how much importance (weight) it should give to each training instance. The weights are modified based on the training performance of the model on the meta-graph. In case the model is ignoring the minority classes (having low performance on the meta-graph), it will increase the weight of those hard to learn minority examples in the main training process.

This meta-graph $G^{meta} = \langle V^{meta}, E^{meta}\rangle$ is constructed using 10\% of the nodes along with their associated edges from the training graph $G \langle V, E \rangle$. Unlike the standard GNN setup, we have two losses here as follows.
\begin{compactenum}
    \item [(i)] \textit{The main loss} (task-specific): $\mathcal{L}_{\text{main}}(\theta) = \frac{1}{N} \sum_{i=1}^N \ell(f_\theta(x_i), y_i)$ where $(x_i,y_i)$ are training samples, i.e., $x_i=p_i$ and $y_i \in \{\textrm{IP, OP}\}$, $f_\theta$ is the GNN model, and $\ell$ is the prediction loss.
    \item [(ii)] The \textit{meta-graph loss} (regularizer): $\mathcal{L}_{\text{meta}}(\theta) = g(\theta, G^{meta})$.
\end{compactenum}
The key idea is that the meta-graph loss is not simply added. Instead, it works through \textit{perturbations of the weights}. The key steps can be enumerated as follows.  
\begin{compactenum}
    \item Compute a candidate weight update from the main loss:
   $
    \theta' = \theta - \eta \nabla_\theta \mathcal{L}_{\text{main}}(\theta),
    $
    where $\eta$ is the learning rate.
    \item Evaluate the meta-graph loss at this perturbed weight $\theta'$:
    $
    \mathcal{L}_{\text{meta}}(\theta') = g(f_{\theta'}, G^{meta}).
    $
    \item Use this meta-loss to refine the gradient update.  
    The total gradient becomes:
    $
    \nabla_\theta \mathcal{L}_{\text{total}} \approx \nabla_\theta \mathcal{L}_{\text{main}}(\theta) + \lambda \, \nabla_\theta \mathcal{L}_{\text{meta}}(\theta'),
    $
    where $\theta'$ reflects how the main loss update affects the meta-graph consistency. 
\end{compactenum}
\noindent\textbf{\textit{Sampling the meta-graph}} ($G^{meta}$): We introduce a novel method to sample $G^{meta}$ from $G$ in such a way that it retains the structural and semantic properties of $G$. We model the sampling of the nodes in $G^{meta}$ as a genetic algorithm problem where the fitness function is designed to ensure that the structural and semantic properties of $G$ are (largely) retained by the resultant $G^{meta}$. The different components of the fitness function are described below.
\begin{compactenum}
    \item \textit{Structural property}: We capture the structural properties using the following metrics.
    \begin{compactenum}
        \item The average degree of $G$ and $G^{meta}$ should be as close as possible: \footnotesize\begin{align} 
f_{\text{deg}} = \frac{\frac{1}{|V^{meta}|} \sum_{j \in V^{meta}} \deg(j)}{\frac{1}{|V|} \sum_{i \in V} \deg(i)}
\end{align}\normalsize
\item  The average \textit{clustering coefficient} (i.e., the extent of `cliquishness') of $G$ and $G^{meta}$ should be as close as possible: \footnotesize\begin{align}
f_{\text{clust}} = \frac{CC_{G^{meta}}}{CC_G}
\end{align}\normalsize
\item The \textit{assortativity} (i.e., the extent of homophily) of $G$ and $G^{meta}$ should be as close as possible: \footnotesize\begin{align}
f_{\text{assort}} = \frac{\rho_{G^{meta}}}{\rho_G}
\end{align}\normalsize
\item The number of communities obtained by clustering (using the method outlined in~\cite{communitydetection}) $G$ ($\mathcal{M}_G$) and $G^{meta}$ ($\mathcal{M}_{G^{meta}}$) should be as close as possible: \footnotesize\begin{align}
f_{\text{comm}} = \frac{\mathcal{M}_{G^{meta}}}{\mathcal{M}_G} \end{align}\normalsize
\end{compactenum}
Overall, $f_{\text{struct}} = f_{\text{deg}} + f_{\text{comm}} + f_{\text{clust}} + f_{\text{assort}}$ represents the structural component of the fitness function.
\item \textit{Semantic property}: We capture the semantic properties using the following metrics.
\begin{compactenum}
\item We compute the variance in each vector entry across all the nodes. The sum of the variances ($\sigma^2$) for all the nodes in $G$ and $G^{meta}$ should be as close as possible:\footnotesize
\begin{align}
f_{\text{var}} = \frac{\sum_{j} \sigma^2(p_{k}^{j}|p_k \in G^{meta})}{\sum_{j} \sigma^2(p_{i}^{j}|p_i \in G)}
\end{align}\normalsize
where $p_{k}^{j}$ is the $j^\textrm{th}$ entry in a patient node $p_{k} \in G^{meta}$ and $p_{i}^{j}$ is the $j^\textrm{th}$ entry in a patient node $p_{i} \in G$.
\item The tree norm \cite{treenorm2025} of a graph ($\left\| G \right\|$) is equivalent to a weighted sum of the number of vertices in the computation trees up to depth $L$. We hypothesize that $f_\text{tn} = \left\| G \right\|_w^L - \left\| G^{meta} \right\|_w^L$\ should be small. The weight $w$ for each depth $l=\{1, \dots, L\}$ is defined as $w_l = \lambda^{l-1}$ where, $\lambda = \exp(-\alpha).\tilde{d_v}$ and $\alpha=1$. The depth $L_v$ for each node $v$ is also dynamically set with a lower $L$ for higher-degree nodes and vice versa. Thus the depth $L_v$ for a node $v$ is computed as $L_v = \lfloor{L_{max}-(L_{max}-L_{min}).\tilde{d_v}}\rfloor$ where $L_{max}$ and $L_{min}$ are the maximum and minimum depths of the tree respectively and $\tilde{d_v}$ is the normalized degree of the node $v$. Mathematically, $\tilde{d_v} = \frac{d_v-d_{min}}{d_{max}-d_{min}}$
where $d_{max}$, $d_{min}$, are respectively the maximum and minimum node degrees in the graph and $d_v$ is the degree of the node $v$. 
\end{compactenum}
The overall semantic fitness is therefore given by $f_\text{sem}=f_\text{var} + f_\text{tn}$.
\end{compactenum}
The total fitness is expressed as $f_\text{total}=f_\text{struct}+f_\text{sem}$. We obtain the best fit $G^{meta}$ using genetic algorithm \cite{GA} with $f_\text{total}$ as the fitness function.
In summary we have the original unbalanced training graph $G$ and a balanced class sub-graph, $G^{meta}$. Now, a GNN is learned and losses are propagated as disscussed above.
\subsection{Trainable base embedding}
In this setup, rather than keeping \fb{} frozen, we make it trainable. Instead of starting with a vanilla sentence transformer, we use the \textsc{GatorTron} language model embeddings (LME)~\cite{gatortron} as it is already pre-trained on biomedical text and, as we shall see later, performs best for our locus-of-care prediction task. The patient notes are initially encoded as 1024 dimensional \textsc{GatorTron} embeddings, concatented with the rest of the features and then input to the GNN framework as earlier. However, the loss is propagated not only to the GNN but also to the top two layers of \textsc{GatorTron} thus adjusting the \fb{} in addition to the parameters of GNN (see Figure~\ref{fig:frmk}).

\if{0}\noindent\textit{3. Tree Norm: }
Tree Movers' Distance \cite{tmd} measures graph distance with fixed depth (typically $\mathcal{L}$ = 4) via optimal transport \cite{ot2019} mechanism. However, this has a time complexity of $\mathcal{O(V}^4)$ where $\mathcal{V}$ is the number of nodes in a graph. Based on TMD, 

\subham{Do we need to include these equations?}
The test compares the goodness of fit of the data with two competing hypotheses \cite{lrt} mentioned below:
\begin{compactenum}    
    \item Null Hypothesis ($H_0$): The occurrence of the ngram is independent of the patient class. The probability of observing the ngram is the same for both IP and OP groups.
    \item Alternative Hypothesis ($H_1$):  The probabilities of the ngram occurring differ between IP and OP groups.
\end{compactenum}
\begin{align}
ll_{\text{null}} 
&= n_{11} \log(p_{11\_\text{null}}) + n_{12} \log(p_{12\_\text{null}}) \notag\\
&\quad + n_{21} \log(p_{11\_\text{null}}) + n_{22} \log(p_{12\_\text{null}}) \\
p_{11\_\text{null}} 
&= \frac{n_{11} + n_{21}}{n}, \quad
p_{12\_\text{null}} = \frac{n_{12} + n_{22}}{n}
\end{align}
\begin{align}
ll_{\text{alt}} 
&= n_{11} \log(p_{11\_\text{alt}}) + n_{12} \log(p_{12\_\text{alt}}) \notag\\
&\quad + n_{21} \log(p_{21\_\text{alt}}) + n_{22} \log(p_{22\_\text{alt}}) \\
p_{11\_\text{alt}} 
&= \frac{n_{11}}{n_{11} + n_{12}}, \quad
p_{12\_\text{alt}} = \frac{n_{12}}{n_{11} + n_{12}}, \notag\\
p_{21\_\text{alt}} 
&= \frac{n_{21}}{n_{21} + n_{22}}, \quad
p_{22\_\text{alt}} = \frac{n_{22}}{n_{21} + n_{22}} \\
n 
&= n_{11} + n_{12} + n_{21} + n_{22}
\end{align}
Here $n_{11}$, $n_{12}$, $n_{21}$, and $n_{22}$ are the respective elements in the 2x2 contingency table \footnote{\url{https://en.wikipedia.org/wiki/Contingency_table}}.

The likelihood ratio test statistic (\textit{lrts}) is calculated as \textit{lrts = -2 * ($ll_{alt}$ - $ll_{null}$)}, where $ll_{alt}$ and $ll_{null}$ are the log-likelihood values of the data under the alternative and null hypotheses, respectively. This lrts statistic aligns with a Chi-Squared ($\chi^2$) distribution with one degree of freedom. We only consider ngrams which are 1\% statistically significant.

\noindent\textit{Meta-graph construction}: We employ genetic algorithm to select meta nodes based on the scores of the fitness function after crossover and mutation. The fitness function is designed such that it captures similar features of the original training graph. The fitness function is represented as a three-fold heuristic optimization problem.

\noindent\textit{1. Structural diversity: } 
This is represented by four complementary structural coverage metrics: degree distribution coverage, community coverage, clustering coefficient, and assortativity.
The mean node degree in the meta-graph $G^{meta}$ is compared against the mean degree of the full graph, $G$. Let deg(i) denote the degree of node $i$. This degree distribution coverage score is expressed as:
\begin{align}
f_{\text{deg}} = \frac{\frac{1}{|G^{meta}|} \sum_{i \in G^{meta}} \deg(i)}{\frac{1}{|G|} \sum_{i \in G} \deg(i)}
\end{align}
The community coverage \cite{communitydetection} and clustering coefficient \cite{clusteringcoeffandcomm} measures the tendency of nodes to form tightly connected groups. The former indicates how tightly these groups are connected, while the latter quantifies the density of connections within the neighborhood of a node. The community coverage and clustering coefficient metric are mathematically represented as: 
\begin{align}
f_{\text{comm}} = \frac{Comm_{G^{meta}}}{Comm_G},  
f_{\text{clust}} = \frac{CC_{G^{meta}}}{CC_G}
\end{align}
respectively.
Assortativity quantifies the correlation between degrees of connected nodes, capturing whether high-degree nodes tend to connect with other high-degree nodes. The ratio of assortativity coefficients between a subgraph and the full graph reveals if the relationships within the meta-graph are representative of the network as a whole.

To this end, we have the final equation combining equations (6), (7), and (8) to capture the structural diversity of the graph, indicated as, .

\noindent\textit{2. Semantic Variance: }
Semantic Variance refers to the degree of diversity in the feature representations (embeddings) of the selected nodes. The ratio of embeddings of selected meta-nodes to that of training nodes represented as $f_{\text{semantic}} = \frac{V_{G^{meta}}}{V_G}$, captures the essence of how varied the collective semantic embeddings of meta-nodes is to that of original nodes. 
Here, $V_{G^{meta}}$ indicates the semantic variance of the meta-graph expressed as $V_{G^{meta}} = \sum_{j=1}^{d} \text{Var}(x_{ij} | i \in {G^{meta}})$ and $V_G$ denotes the semantic variance of full training graph mathematically represented as, $V_{\text{total}} = \sum_{j=1}^{d} \text{Var}(x_{1j}, x_{2j}, \ldots, x_{Nj})$ where $d$ and $x_{ij}$ denote the embedding dimension and $j^{th}$ feature of node $i$ respectively.

\fi
\section{Experimental Setup}
\label{sec:experiment}
\noindent\textbf{Variants of \ours{}}: For \ours{}, we present results for four GNN variants -- \grgcn{} (\textsc{GCN} as the GNN), \grgsg{} (\textsc{GraphSage} as the GNN), \grgat{} (\textsc{GAT} as the GNN), \grgt{} (\textsc{Graph transformers} as the GNN), and one variant with trainable \fb{} -- \grtron{} (\textsc{GraphSage} as the GNN since, as we shall see, it performs best among the GNN variants). \\
\noindent\textbf{Hyperparameters}:
The GNN in \ours{} consists of a single fully connected linear layer, followed by two GNN layers. The hidden dimensions ($hd$) were tuned within the range of 8 to 256 units, with ReLU activation functions. In addition, the learning rate ($lr$) is set to be optimized in the range of (1e-5 to 1e-2). The classification output came from a log-softmax layer where dropout-based regularization was finally applied for binary prediction. All these hyperparameters are tuned using Optuna\footnote{\url{https://optuna.org/}}. We run Optuna for 100 trials for each of the experiments. The best hyperparameter values obtained using Optuna are $hd = 191$, $lr = 0.0048$ and $lr_{meta} = 0.00287$ respectively. For the trainable \fb{}, we used AdamW optimizer for 10 epochs with batch size of 1; all the other hyperparameters such as weight decay, dropout, etc. are defaulted to the \textsc{GatorTron} base model.\\
\noindent\textbf{Baselines}:
We compare the performance of our model against \textit{five} categories of baselines, including traditional ML algorithms, deep learning, transformer based language model, finetuned medical LLMs and a bunch of recent techniques. 
Traditional models include logistic regression (LR) and \textsc{SVM}. The deep learning models \textsc{Bi-LSTM} and \textsc{BERT-ft} are finetuned on the training set to compare with \ours{}.
We also compare the performance of \ours{} with three of the recent works -- ~\cite{deepj,patemb,graphgpt}. Baselines using transformer architectures in a zero-shot setting include \textsc{Qwen32}~\shortcite{web5}, a multilingual reasoning LLM developed for a variety of complex reasoning tasks, \textsc{Deepseek-R1}~\shortcite{web6}, an advanced generative model designed for retrieval and logical reasoning, \textsc{GPT-oss} \shortcite{web1}, an open-weight 120b reasoning model that achieves competitive scores in medical tasks, \textsc{GatorTron}\footnote{\url{https://huggingface.co/UFNLP/gatortron-base-2k}}, a large clinical language model trained from scratch on $>82$ billion words of de-identified clinical text, and \textsc{MedGemma}\footnote{\url{https://huggingface.co/google/medgemma-27b-text-it}}, a fine-tuned vision-language foundation model based on Google's open model Gemma \cite{gemma} trained on medical data. Further, \textsc{GatorTron-ft} and \textsc{MedGemma-ft} represents the fine-tuned versions of their base model on our data. For detailed fine-tuning configuration for both of these models see \textit{supplementary materials}.

\noindent\textbf{Evaluation metrics}:
We evaluate \ours{} based on the classwise precision, recall, and F1-score. In addition, we also report accuracy and AUROC.

\section{Results and Discussion}
\label{sec:results}
\begin{table*}
\footnotesize
    \centering
    \begin{tabular}{lcccccccc}
    \toprule
         \textbf{Models} & \textbf{IP-PR} & \textbf{IP-R} & \textbf{IP-$\mathbf{F_1}$} & \textbf{OP-PR} & \textbf{OP-R} & \textbf{OP-$\mathbf{F_1}$} & \textbf{Acc} & \textbf{AUROC} \\
         \midrule
         LR & 0.47 & 0.56 & 0.51 & 0.73 & 0.65 & 0.69 & 0.62 & 0.65 \\
         SVM & 0.51 & 0.50 & 0.51 & 0.73 & 0.74 & 0.74 & 0.67 & 0.66 \\
         \midrule
         \textsc{Bi-LSTM} & 0.46 & 0.43 & 0.45 & 0.70 & 0.72 & 0.71 & 0.62 & 0.61 \\
         \textsc{BERT-ft} & 0.72 & 0.48 & 0.58 & 0.76 & 0.90 & 0.82 &0.75 & 0.79 \\
         \midrule
         \textsc{DeepJ} & 0.40 & 0.56 & 0.47 & 0.69 & 0.55 & 0.61 & 0.55 & - \\
         \textsc{GraphGPT} & 0.56 & 0.55 & 0.56 & 0.76 & 0.70 & 0.73 & 0.68 & - \\
         \textsc{PatEmb} & 0.59 & 0.67 & 0.63 & 0.81 & 0.75 & 0.78 & 0.72 & - \\
         \midrule
         \textsc{Qwen32} & 0.51 & 0.56 & 0.53 & 0.74 & 0.70 & 0.72 & 0.66 & - \\
         \textsc{Deepseek-R1} & 0.53 & 0.56 & 0.55 & 0.75 & 0.73 & 0.74 & 0.68 & - \\
         \textsc{GPT-oss} & 0.51 & 0.57 & 0.54 & 0.75 & 0.69 & 0.72 & 0.65 & - \\
         \textsc{GatorTron} & 0.56 & 0.79 & 0.66 & 0.85 & 0.66 & 0.75 & 0.71 & - \\
         \textsc{MedGemma} & 0.41 & 0.93 & 0.57 & 0.88 & 0.29 & 0.43 & 0.51 & - \\
         \midrule
         \textsc{GatorTron-ft} & 0.67 & \sethlcolor{yellow}\hl{0.85} & 0.74 & \sethlcolor{yellow}\hl{0.90} & 0.77 & 0.83 & 0.79 & - \\
         \textsc{MedGemma-ft} & \sethlcolor{yellow}\hl{0.88} & 0.77 & \sethlcolor{yellow}\hl{0.82} & 0.88 & \sethlcolor{yellow}\hl{0.94} & \sethlcolor{yellow}\hl{0.91} & \sethlcolor{yellow}\hl{0.88} & - \\
         \midrule
         \grgcn{} & 0.63 & 0.48 & 0.55$^*$ & 0.75 & 0.85 & 0.80$^*$ & 0.72 & 0.73 \\
         \grgat{} & 0.74 & 0.47 & 0.57$^*$ & 0.76 & 0.91 & 0.83$^*$ & 0.75 & 0.74 \\
         \grgt{} & 0.82 & 0.67 & 0.73$^{**}$ & 0.83 & 0.91 & 0.87$^{**}$ & 0.83 & 0.88 \\
         \grgsg{} & 0.85 & 0.64 & 0.74$^{**}$ & 0.83 & \sethlcolor{yellow}\hl{0.94} & 0.88$^{**}$ & 0.84 & \sethlcolor{yellow}\hl{0.89} \\
         \textsc{\grtron{}} & \sethlcolor{lightgreen}\hl{0.97} & \sethlcolor{lightgreen}\hl{0.93} & \sethlcolor{lightgreen}\hl{0.95$^{**}$} & \sethlcolor{lightgreen}\hl{0.96} & \sethlcolor{lightgreen}\hl{0.98} & \sethlcolor{lightgreen}\hl{0.97$^{**}$} & \sethlcolor{lightgreen}\hl{0.96} & \sethlcolor{lightgreen}\hl{0.99} \\
    \bottomrule
    \end{tabular}
  \caption{\label{tab:ipop}
    \footnotesize Performance comparison of all the variants of \ours{} with the competing baselines. Best and second best results are highlighted in green and yellow respectively. PR: Precision, R: Recall, Acc: Accuracy. We report the Friedman omnibus test to check statistical significance of \ours{} models. * indicates p-value $< 0.01$, while ** indicates p-value $< 0.001$.}
\end{table*}

\begin{table}[h]
\centering
\footnotesize
\begin{tabular}{lccc}
\toprule
\textbf{Embeddings} & \textbf{IP-$\mathbf{F}_1$} & \textbf{OP-$\mathbf{F}_1$} & \textbf{AUROC} \\
\midrule
 \fb{} & 0.51 & 0.77 & 0.69 \\
 \hdashline
 \hspace{1mm}+\fe{} & 0.53 & 0.76 & 0.71 \\
 \hspace{1mm}+\fl{} & 0.55 & 0.84 & 0.83 \\
 \hspace{1mm}+\fe{}+\fl{} & 0.57 & 0.84 & 0.84 \\
 \hspace{1mm}+\fe{}+\fl{}+\fr{} & 0.74 & 0.88 & 0.89 \\
\bottomrule
\end{tabular}
\caption{\footnotesize Ablation study on node features using \grgsg{}.}
\label{tab:encoding_ablation}
\end{table}

In this section, we systematically evaluate the \ours{} framework and compare the results with the baselines.\\ 
\subsection{Performance evaluation}
Table~\ref{tab:ipop} compares the results of \ours{} with all the baselines using the different evaluation metrics. The \grtron{} variant by far outperforms all the baselines by a large margin. LR and SVM achieve moderate F1 scores with AUROC of \textasciitilde{}0.66, highlighting their limitations when handling complex high-dimensional data. The reasoning-based LLMs in a zero-shot setting show slight improvement when compared with traditional ML algorithms. Although \textsc{BERT-ft} finetuned on our data encodes the text better, it fails to capture information from neighbouring nodes, leading to weaker inpatient (IP) predictions. Among the \ours{} variants, \grtron{} performs the best followed by \grgsg{}. \grtron{} also outperforms the best SOTA medical-LLMs (\textsc{MedGemma-ft}) finetuned on our dataset by 15.8\% in terms of IP F1 score. In fact, \grtron{} achieves near perfect scores demonstrating the suitability of \ours{} in real-world clinical decision support, where missing subtle patterns can lead to critical misclassifications.
\subsection{Ablation experiments} 
Here, we briefly report the results of the ablation experiments on binary classification task by sequentially adding the node features one by one. Table~\ref{tab:encoding_ablation} notes the classwise F1 scores demonstrating that inclusion of \fe{}, \fl{} and \fr{} systematically improves the overall performance. In addition, we conduct ablations of meta-graph and the heuristics involved in the fitness function of the genetic algorithm to obtain unbiased meta-graph. We observe that the F1 scores of the minority class are 0.67 when we omit the structural properties, and 0.65 when we omit the semantic properties. This suggests that each of the heuristics has a significant contribution towards the sampling of $G^{meta}$. 

\subsection{Discussion}
\begin{figure}[t]
  \includegraphics[width=\linewidth]{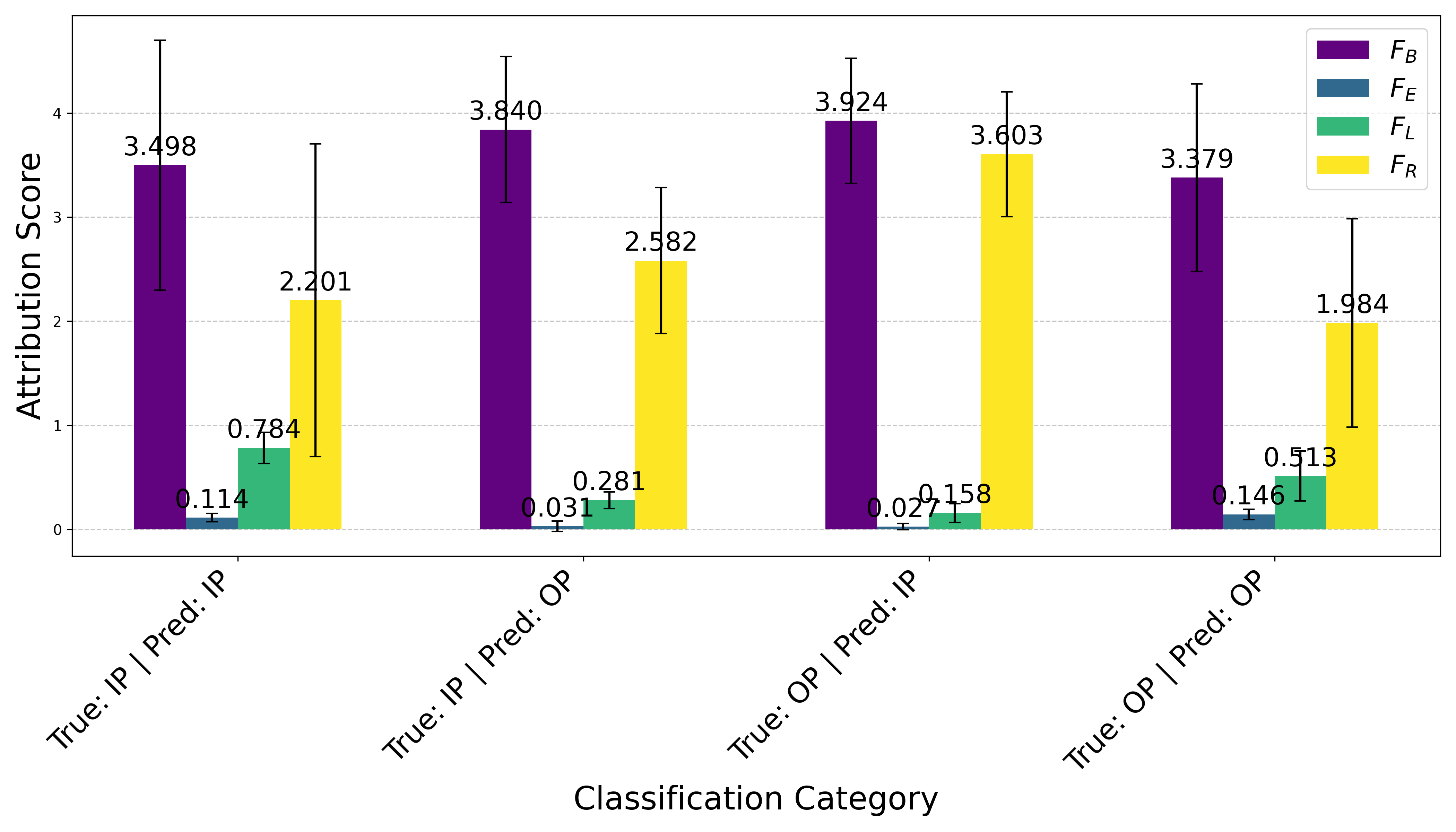}
  \caption{\footnotesize Category-wise contribution of each feature (\fb{}, \fe{}, \fl{}, and \fr{}) on the final outcome of \grtron{}. $x$-axis denotes the classification categories, while $y$-axis represents the mean attribution scores for each category.}
  \label{fig:feature_contri}
\end{figure}

\noindent\textbf{Number of parameters}: The number of trainable parameters for \ours{} is far less compared to the state-of-the-art biomedical LLMs. The total number of parameters for \grtron{} is only 14M compared to 345M and 27B for \textsc{GatorTron} and \textsc{MedGemma} respectively. These models are far behind \grtron{} in their zero-shot setting; while finetuning helps, it is not able to reach the performance level of \grtron{}. Interestingly, \grgsg{} which is the best among the graph-only versions of \ours{} has just 2M trainable parameters but reaches very close to the performance of \textsc{GatorTron-ft} and \textsc{MedGemma-ft}. This indicates that the graph formulation plays an important part in enhanching the performance of \ours{}. This also simulates real-world scenario as clinicians do not treat patients in isolation; they rather mentally compare similar patients from their experience, drawing correlation among patient cases, treatments, and outcomes. In fact, there are several earlier studies which demonstrate that the usage of patient similarity graph in medical applications is comparatively a better design choice~\cite{PSN2018,PSNheart2024,GNNEHR2024,PSNincancer}.  \\
\noindent\textbf{Choice of edge threshold}: We selected the threshold for cosine similarity $\tau = 0.80$ based on emperical observations. Slight changes to this value does not affect the result indicating that the trends are robust. However using very low thresholds results in denser graph that contain more aggregated information, but at the same time introduce noise by connecting irrelevant patient pairs; using even higher thresholds results in sparse graph of high confidence but at the same time introduces the risk of fragmenting the information and poor information flow. With addiction patient populations, setting $\tau$ close to 0.80 translates clinically meaningful information into practice.\\
\noindent\textbf{Importance of meta-learning}: Training \grtron{} without the meta-learning component results in a drop of 7\% in overall performance. This shows that the dynamic class-reweighting introduced through meta-learning actually impacts the performance of the model. We further checked the performance of the different variants of \ours{} by increasing the size of meta-graph. However, there was negligible change in performance while the training time increased significantly (e.g., using 25\% instead of 10\% meta nodes increases the training time by 14\%). \\
\noindent\textbf{Impact of the features}: In order to understand the impact of the different node features, i.e., \fb{}, \fe{}, \fl{} and \fr{} we compute attribution scores for each of them using the integrated gradient method. These scores indicate how much importance a model gives to each of these features while making a prediction. We use \textsc{Captum}\footnote{\url{https://captum.ai/docs/extension/integrated_gradients}}~\cite{captum} to compute the attribution scores associated with each feature for the best model -- \grtron{}. Overall, we find that \fb{} and \fr{} contribute the most to the final predicted outcome with respective attribution score shares of 54.09\% and 38.31\%. \fl{} and \fe{} contribute only 6.4\% and 1.2\% respectively. To deep dive, we plot in Figure~\ref{fig:feature_contri} the contributions of these feature embeddings for all four categories -- correct classification, where (true\_label-predicted\_label) are the same (i.e., IP-IP and OP-OP) and misclassifications, where (true\_label-predicted\_label) differ (i.e., IP-OP and OP-IP). The base embeddings \fb{} consistently achieves higher attribution scores as it integrates LME and the network neighbourhood information, capturing contextual and relational knowledge. \fr{} is the next most important predictor; manual inspection by psychiatrists shows that the high attribution score for the OP-IP misclassification is mainly due to the confusing mixed signals in a note (e.g., `complaints of seizure' and `reported abstinence from alcohol' contradict each other). Interestingly, although the absolute attribution scores for \fl{} are low across all the categories, it has a marked stronger influence on the correct prediction categories (especially, the IP-IP category, which is also the hardest).

\section{Conclusion}
\label{sec:conclusion}
The \ours{} framework proposed in this work addresses the challenge of determining the locus of care in imbalanced addiction data. Our work contributes to both methodological innovations and practical insights at the intersection of artificial intelligence and addiction medicine. In terms of methods, it intelligently encodes patient notes into a high-dimensional latent space, introduces meta-learning to handle data imbalance, and jointly fine-tunes the GNN and the base embeddings to by far outperform all the state-of-the-art baselines. The comprehensive evaluation establishes the superiority of the proposed method. 
Overall, this work presents a promising and novel contribution to advancing clinical decision-making in addiction treatment.

\bibliographystyle{named}
\bibliography{ijcai26}

@misc{web1,
  title = {GPT-OSS},
  author = {OpenAI},
  year = {2025},
  url = {https://openrouter.ai/openai/gpt-oss-120b},
}

@misc{web4,
  title = {LLaMa-3-Instruct},
  author = {Llama-Team},
  year = {2025},
  url = {https://huggingface.co/meta-llama/Llama-3.2-3B-Instruct},
}

@misc{web5,
  title = {Qwen3-32b},
  author = {Qwen-Team},
  year = {2025},
  url = {https://openrouter.ai/qwen/qwen3-32b},
}

@misc{web6,
  title = {Deepseek-R1},
  author = {Deepseek},
  year = {2025},
  url = {https://openrouter.ai/deepseek/deepseek-r1},
}

@article{Goodson2014AWS,
  author    = {Goodson, Christopher M. and Clark, Benjamin J. and Douglas, Ivor S.},
  title     = {Predictors of Severe Alcohol Withdrawal Syndrome: A Systematic Review and Meta-Analysis},
  journal   = {Alcoholism: Clinical and Experimental Research},
  year      = {2014},
  volume    = {38},
  number    = {10},
  pages     = {2664--2677},
}

@article{empath,
author = {Fast, Ethan and Chen, Binbin and Bernstein, Michael},
year = {2016},
month = {02},
pages = {},
title = {Empath: Understanding Topic Signals in Large-Scale Text},
doi = {10.1145/2858036.2858535}
}

@book{llrtfrombook,
  address = {Cambridge, Massachusetts},
  author = {Manning, Christopher D. and Sch{\"u}tze, Hinrich},
  publisher = {The {MIT} Press},
  title = {Foundations of Statistical Natural Language Processing},
  url = {http://nlp.stanford.edu/fsnlp/},
  year = 1999
}

@article{lrt,
 ISSN = {00905364, 21688966},
 URL = {http://www.jstor.org/stable/23566539},
 author = {Tiefeng Jiang and Fan Yang},
 journal = {The Annals of Statistics},
 number = {4},
 pages = {2029--2074},
 publisher = {Institute of Mathematical Statistics},
 title = {CENTRAL LIMIT THEOREMS FOR CLASSICAL LIKELIHOOD RATIO TESTS FOR HIGH-DIMENSIONAL NORMAL DISTRIBUTIONS},
 urldate = {2025-08-13},
 volume = {41},
 year = {2013}
}

@article{metaGCN,
   title={Meta-GCN: A Dynamically Weighted Loss Minimization Method for Dealing with the Data Imbalance in Graph Neural Networks},
   url={http://dx.doi.org/10.21428/594757DB.0041F830},
   DOI={10.21428/594757db.0041f830},
   journal={Proceedings of the Canadian Conference on Artificial Intelligence},
   publisher={PubPub},
   author={Mohammadizadeh, Mahdi and Mozhdehi, Arash and Ioannou, Yani and Wang, Xin},
   year={2023},
   month=jun }

@article{communitydetection,
   title={Finding community structure in very large networks},
   volume={70},
   ISSN={1550-2376},
   url={http://dx.doi.org/10.1103/PhysRevE.70.066111},
   DOI={10.1103/physreve.70.066111},
   number={6},
   journal={Physical Review E},
   publisher={American Physical Society (APS)},
   author={Clauset, Aaron and Newman, M. E. J. and Moore, Cristopher},
   year={2004},
   month=dec }

@article{clusteringcoeffandcomm,
title = {Clustering coefficient and community structure of bipartite networks},
journal = {Physica A: Statistical Mechanics and its Applications},
volume = {387},
number = {27},
pages = {6869-6875},
year = {2008},
issn = {0378-4371},
doi = {https://doi.org/10.1016/j.physa.2008.09.006},
url = {https://www.sciencedirect.com/science/article/pii/S0378437108007796},
author = {Peng Zhang and Jinliang Wang and Xiaojia Li and Menghui Li and Zengru Di and Ying Fan}
}

@article{assortativity,
author = {Thedchanamoorthy, Gnanakumar and Piraveenan, Mahendra and Kasthuriratna, Dharshana and Senanayake, Upul},
year = {2014},
month = {12},
pages = {},
title = {Node Assortativity in Complex Networks: An Alternative Approach},
volume = {29},
journal = {Procedia Computer Science},
doi = {10.1016/j.procs.2014.05.229}
}

@article{SUDML,
author = {Gail K. Strickler and Sharon Reif and Constance M. Horgan and Andrea Acevedo},
title = {The Relationship Between Substance Abuse Performance Measures and Mutual-Help Group Participation after Treatment},
journal = {Alcoholism Treatment Quarterly},
volume = {30},
number = {2},
pages = {190--210},
year = {2012},
publisher = {Taylor \& Francis},
doi = {10.1080/07347324.2012.663305},
    note ={PMID: 22879689},
URL = {https://doi.org/10.1080/07347324.2012.663305}
}

@article{GA,
  author    = {Goldberg, D.E. and Holland, J.H.},
  title     = {Genetic Algorithms and Machine Learning},
  journal   = {Machine Learning},
  volume    = {3},
  number    = {2},
  pages     = {95--99},
  year      = {1988},
  publisher = {Kluwer Academic Publishers},
  doi       = {10.1023/A:1022602019183}
}

@article{sud2017use,
  title={Use of a machine learning framework to predict substance use disorder treatment success},
  author={Acion, Laura and Kelmansky, Diana and van der Laan, Mark and Sahker, Ethan and Jones, DeShauna and Arndt, Stephan},
  journal={PloS one},
  volume={12},
  number={4},
  pages={e0175383},
  year={2017},
  publisher={Public Library of Science San Francisco, CA USA}
}

@article{aud2013increased,
  title={Increased risk of alcohol and drug use among children from deployed military families},
  author={Acion, Laura and Ramirez, Marizen R and Jorge, Ricardo E and Arndt, Stephan},
  journal={Addiction},
  volume={108},
  number={8},
  pages={1418--1425},
  year={2013},
  publisher={Wiley Online Library}
}

@article{aud-dss2023,
  author = {Ebrahimi, Ali and Wiil, Uffe Kock and Baskaran, Ruben and Peimankar, Abdolrahman and Andersen, Kjeld and Nielsen, Anette Søgaard},
  title = {AUD-DSS: a decision support system for early detection of patients with alcohol use disorder},
  journal = {BMC Bioinformatics},
  volume = {24},
  number = {1},
  pages = {329},
  year = {2023},
  month = {September},
  doi = {10.1186/s12859-023-05450-6},
  url = {https://doi.org/10.1186/s12859-023-05450-6}
}

@article{MLsurvey,
title = {How machine learning is used to study addiction in digital healthcare: A systematic review},
journal = {International Journal of Information Management Data Insights},
volume = {3},
number = {2},
pages = {100175},
year = {2023},
issn = {2667-0968},
doi = {https://doi.org/10.1016/j.jjimei.2023.100175},
url = {https://www.sciencedirect.com/science/article/pii/S2667096823000228},
author = {Bijoy Chhetri and Lalit Mohan Goyal and Mamta Mittal}
}

@article{SVM2015,
title = {Interpreting support vector machine models for multivariate group wise analysis in neuroimaging},
journal = {Medical Image Analysis},
volume = {24},
number = {1},
pages = {190-204},
year = {2015},
issn = {1361-8415},
doi = {https://doi.org/10.1016/j.media.2015.06.008},
url = {https://www.sciencedirect.com/science/article/pii/S136184151500095X},
author = {Bilwaj Gaonkar and Russell {T. Shinohara} and Christos Davatzikos}
}

@article{weightedPSN2021,
  author = {Lu, Haohui and Uddin, Shahadat},
  title = {A weighted patient network-based framework for predicting chronic diseases using graph neural networks},
  journal = {Scientific Reports},
  volume = {11},
  number = {1},
  pages = {22607},
  year = {2021},
  doi = {10.1038/s41598-021-01964-2},
  url = {https://doi.org/10.1038/s41598-021-01964-2}
}

@article{GNNlit2020,
  title={A comprehensive survey on graph neural networks},
  author={Wu, Zonghan and Pan, Shirui and Chen, Fengwen and Long, Guodong and Zhang, Chengqi and Yu, Philip S},
  journal={IEEE transactions on neural networks and learning systems},
  volume={32},
  number={1},
  pages={4--24},
  year={2020},
  publisher={IEEE}
}

@article{preciseADR,
  author = {Gao, Yang and Zhang, Xiang and Sun, Zhongquan and Chandak, Payal and Bu, Jiajun and Wang, Haishuai},
  title = {Precision Adverse Drug Reactions Prediction with Heterogeneous Graph Neural Network},
  journal = {Advanced Science},
  year = {2024},
  doi = {10.1002/advs.202404671},
  url = {https://doi.org/10.1002/advs.202404671}
}

@article{lighted,
title = {An integrated LSTM-HeteroRGNN model for interpretable opioid overdose risk prediction},
journal = {Artificial Intelligence in Medicine},
volume = {135},
pages = {102439},
year = {2023},
issn = {0933-3657},
doi = {https://doi.org/10.1016/j.artmed.2022.102439},
url = {https://www.sciencedirect.com/science/article/pii/S0933365722001919},
author = {Xinyu Dong and Rachel Wong and Weimin Lyu and Kayley Abell-Hart and Jianyuan Deng and Yinan Liu and Janos G. Hajagos and Richard N. Rosenthal and Chao Chen and Fusheng Wang}
}

@article{PSN2018,
title = {Patient Similarity Networks for Precision Medicine},
journal = {Journal of Molecular Biology},
volume = {430},
number = {18, Part A},
pages = {2924-2938},
year = {2018},
note = {Theory and Application of Network Biology Toward Precision Medicine},
issn = {0022-2836},
doi = {https://doi.org/10.1016/j.jmb.2018.05.037},
url = {https://www.sciencedirect.com/science/article/pii/S0022283618305321},
author = {Shraddha Pai and Gary D. Bader}
}

@article{netDx,
author = {Pai, Shraddha and Hui, Shirley and Isserlin, Ruth and Shah, Muhammad A and Kaka, Hussam and Bader, Gary D},
title = {netDx: interpretable patient classification using integrated patient similarity networks},
journal = {Molecular Systems Biology},
volume = {15},
number = {3},
pages = {e8497},
doi = {https://doi.org/10.15252/msb.20188497},
url = {https://www.embopress.org/doi/abs/10.15252/msb.20188497},
eprint = {https://www.embopress.org/doi/pdf/10.15252/msb.20188497},
year = {2019}
}

@misc{llmSUDsocialmedia,
      title={Leveraging Large Language Models for Multi-Class and Multi-Label Detection of Drug Use and Overdose Symptoms on Social Media}, 
      author={Muhammad Ahmad and Fida Ullah and Muhammad Usman and Umyh Habiba and ldar Batyrshin and Grigori Sidorov},
      year={2025},
      eprint={2504.12355},
      archivePrefix={arXiv},
      primaryClass={cs.CL},
      url={https://arxiv.org/abs/2504.12355}, 
}

@article{llmSUD,
  title={Stigmatizing Language in Large Language Models for Alcohol and Substance Use Disorders: A Multimodel Evaluation and Prompt Engineering Approach},
  author={Wang, Yichen and Hsu, Kelly and Brokus, Christopher and Huang, Yuting and Ufere, Nneka and Wakeman, Sarah and Zou, James and Zhang, Wei},
  journal={Journal of Addiction Medicine},
  doi={ 10.1097/ADM.0000000000001536},
  pages={10--1097},
  year={2024},
  publisher={LWW}
}

@article{adaptivetreatmentQlearning,
  author = {Nahum-Shani, Inbal and Ertefaie, Ashkan and Lu, Xi and Lynch, Kevin G. and McKay, James R. and Oslin, David W. and Almirall, Daniel},
  title = {A SMART data analysis method for constructing adaptive treatment strategies for substance use disorders},
  journal = {Addiction},
  volume = {112},
  number = {5},
  pages = {901--909},
  year = {2017},
  doi = {10.1111/add.13743}
}

@article{cdss,
  author = {Sutton, Reed T. and Pincock, David and Baumgart, Daniel C. and Sadowski, Daniel C. and Fedorak, Richard N. and Kroeker, Karen I.},
  title = {An overview of clinical decision support systems: benefits, risks, and strategies for success},
  journal = {npj Digital Medicine},
  volume = {3},
  number = {1},
  pages = {17},
  year = {2020},
  doi = {10.1038/s41746-020-0221-y},
  url = {https://doi.org/10.1038/s41746-020-0221-y},
}

@article{Volkow2023,
  author = {Volkow, Nora D. and Blanco, Carlos},
  title = {Substance use disorders: a comprehensive update of classification, epidemiology, neurobiology, clinical aspects, treatment and prevention},
  journal = {World Psychiatry},
  volume = {22},
  number = {2},
  pages = {203--229},
  year = {2023},
  doi = {10.1002/wps.21073}
}

@article{covid19,
    author = {Narasimha, Venkata Lakshmi and Shukla, Lekhansh and Mukherjee, Diptadhi and Menon, Jayakrishnan and Huddar, Sudheendra and Panda, Udit Kumar and Mahadevan, Jayant and Kandasamy, Arun and Chand, Prabhat K and Benegal, Vivek and Murthy, Pratima},
    title = {Complicated Alcohol Withdrawal—An Unintended Consequence of COVID-19 Lockdown},
    journal = {Alcohol and Alcoholism},
    volume = {55},
    number = {4},
    pages = {350-353},
    year = {2020},
    month = {05},
    issn = {0735-0414},
    doi = {10.1093/alcalc/agaa042},
    url = {https://doi.org/10.1093/alcalc/agaa042},
    eprint = {https://academic.oup.com/alcalc/article-pdf/55/4/350/33414426/agaa042.pdf},
}

@article{unknown,
author = {Gururaj, G and Varghese, Mathew and Benegal, Vivek and Rao, Girish and Pathak, Komal and Singh, Lokesh and Mehta, Ritambhara and D, Ram and Shibukumar, Tm and Kokane, Arun and RK, Lenin and BS, Chavan and P, Sharma and C, Ramasubramanian and Dalal, Pronob and Saha, Pranesh and SP, Deuri and Giri, Anjan and AB, Kavishvar and Goyal, Nishant},
year = {2017},
month = {03},
pages = {},
title = {National Mental Health Survey of India, 2015-16 Prevalence, Pattern and Outcomes},
doi ={https://journals.lww.com/indianjpsychiatry/fulltext/2017/59010/national_mental_health_survey_of_india_2015_2016.7.aspx}
}

@article{gaps,
  author = {Lamb S, Greenlick MR, McCarty D},
  title = {Bridging the Gap between Practice and Research: Forging Partnerships with Community-Based Drug and Alcohol Treatment.},
  journal = {National Academies Press (US), Washington (DC)},
  year = {1998},
  howpublished = "\url{https://www.ncbi.nlm.nih.gov/books/NBK230400/}"
}

@misc{gcn,
      title={Semi-Supervised Classification with Graph Convolutional Networks}, 
      author={Thomas N. Kipf and Max Welling},
      year={2017},
      eprint={1609.02907},
      archivePrefix={arXiv},
      primaryClass={cs.LG},
      url={https://arxiv.org/abs/1609.02907}, 
}

@misc{graphsage,
      title={Inductive Representation Learning on Large Graphs}, 
      author={William L. Hamilton and Rex Ying and Jure Leskovec},
      year={2018},
      eprint={1706.02216},
      archivePrefix={arXiv},
      primaryClass={cs.SI},
      url={https://arxiv.org/abs/1706.02216}, 
}

@misc{gat,
      title={Graph Attention Networks}, 
      author={Petar Veličković and Guillem Cucurull and Arantxa Casanova and Adriana Romero and Pietro Liò and Yoshua Bengio},
      year={2018},
      eprint={1710.10903},
      archivePrefix={arXiv},
      primaryClass={stat.ML},
      url={https://arxiv.org/abs/1710.10903}, 
}

@misc{tmd,
      title={Tree Mover's Distance: Bridging Graph Metrics and Stability of Graph Neural Networks}, 
      author={Ching-Yao Chuang and Stefanie Jegelka},
      year={2022},
      eprint={2210.01906},
      archivePrefix={arXiv},
      primaryClass={cs.LG},
      url={https://arxiv.org/abs/2210.01906}, 
}

@misc{treenorm2025,
      title={Subsampling Graphs with GNN Performance Guarantees}, 
      author={Mika Sarkin Jain and Stefanie Jegelka and Ishani Karmarkar and Luana Ruiz and Ellen Vitercik},
      year={2025},
      eprint={2502.16703},
      archivePrefix={arXiv},
      primaryClass={cs.LG},
      url={https://arxiv.org/abs/2502.16703}, 
}

@misc{ot2019,
      title={Optimal Transport for structured data with application on graphs}, 
      author={Titouan Vayer and Laetitia Chapel and Rémi Flamary and Romain Tavenard and Nicolas Courty},
      year={2019},
      eprint={1805.09114},
      archivePrefix={arXiv},
      primaryClass={stat.ML},
      url={https://arxiv.org/abs/1805.09114}, 
}

@book{WHO1993ICD10,
  title = {The ICD-10 Classification of Mental and Behavioural Disorders: Diagnostic Criteria for Research},
  publisher = {World Health Organization},
  address = {Geneva},
  year = {1993},
  author = {{World Health Organization}}
}

@misc{note_correction,
 title={Language Models for standardising clinical notes and information extraction in addiction psychiatry – an empirical study},
 url={osf.io/d5m6e_v1},
 DOI={10.31219/osf.io/d5m6e_v1},
 publisher={OSF Preprints},
 author={Shukla, Lekhansh},
 year={2025},
 month={Feb}
}

@misc{gliner,
      title={GLiNER multi-task: Generalist Lightweight Model for Various Information Extraction Tasks}, 
      author={Ihor Stepanov and Mykhailo Shtopko},
      year={2024},
      eprint={2406.12925},
      archivePrefix={arXiv},
      primaryClass={cs.LG},
      url={https://arxiv.org/abs/2406.12925}, 
}

@article{Goodson2014,
  author = {Goodson, Carrie M and Clark, Brendan J and Douglas, Ivor S},
  title = {Predictors of severe alcohol withdrawal syndrome: a systematic review and meta-analysis},
  journal = {Alcoholism, Clinical and Experimental Research},
  volume = {38},
  number = {10},
  pages = {2664--2677},
  year = {2014},
  doi = {10.1111/acer.12529},
  pmid = {25346507},
  pmc = {PMC6299173}
}

@misc{graphtranformer,
      title={Masked Label Prediction: Unified Message Passing Model for Semi-Supervised Classification}, 
      author={Yunsheng Shi and Zhengjie Huang and Shikun Feng and Hui Zhong and Wenjin Wang and Yu Sun},
      year={2021},
      eprint={2009.03509},
      archivePrefix={arXiv},
      primaryClass={cs.LG},
      url={https://arxiv.org/abs/2009.03509}, 
}

@inproceedings{optuna_2019,
    title={Optuna: A Next-generation Hyperparameter Optimization Framework},
    author={Akiba, Takuya and Sano, Shotaro and Yanase, Toshihiko and Ohta, Takeru and Koyama, Masanori},
    booktitle={Proceedings of the 25th {ACM} {SIGKDD} International Conference on Knowledge Discovery and Data Mining},
    year={2019}
}

@inproceedings{graphgpt,
author = {Tang, Jiabin and Yang, Yuhao and Wei, Wei and Shi, Lei and Su, Lixin and Cheng, Suqi and Yin, Dawei and Huang, Chao},
title = {GraphGPT: Graph Instruction Tuning for Large Language Models},
year = {2024},
isbn = {9798400704314},
publisher = {Association for Computing Machinery},
address = {New York, NY, USA},
url = {https://doi.org/10.1145/3626772.3657775},
doi = {10.1145/3626772.3657775},
booktitle = {Proceedings of the 47th International ACM SIGIR Conference on Research and Development in Information Retrieval},
pages = {491–500},
numpages = {10},
location = {Washington DC, USA},
series = {SIGIR '24}
}

@misc{deepj,
      title={DeepJ: Graph Convolutional Transformers with Differentiable Pooling for Patient Trajectory Modeling}, 
      author={Deyi Li and Zijun Yao and Muxuan Liang and Mei Liu},
      year={2025},
      eprint={2506.15809},
      archivePrefix={arXiv},
      primaryClass={cs.LG},
      url={https://arxiv.org/abs/2506.15809}, 
}

@article{patemb,
  title = {Transformer patient embedding using electronic health records enables patient stratification and progression analysis},
  volume = {8},
  ISSN = {2398-6352},
  url = {http://dx.doi.org/10.1038/s41746-025-01872-z},
  DOI = {10.1038/s41746-025-01872-z},
  number = {1},
  journal = {npj Digital Medicine},
  publisher = {Springer Science and Business Media LLC},
  author = {Xian,  Su and Grabowska,  Monika E. and Kullo,  Iftikhar J. and Luo,  Yuan and Smoller,  Jordan W. and Walunas,  Theresa L. and Wei,  Wei-Qi and Jarvik,  Gail P. and Mooney,  Sean D. and Crosslin,  David R.},
  year = {2025},
  month = aug 
}

@article{gatortron,
  title = {A large language model for electronic health records},
  volume = {5},
  ISSN = {2398-6352},
  url = {http://dx.doi.org/10.1038/s41746-022-00742-2},
  DOI = {10.1038/s41746-022-00742-2},
  number = {1},
  journal = {npj Digital Medicine},
  publisher = {Springer Science and Business Media LLC},
  author = {Yang,  Xi and Chen,  Aokun and PourNejatian,  Nima and Shin,  Hoo Chang and Smith,  Kaleb E. and Parisien,  Christopher and Compas,  Colin and Martin,  Cheryl and Costa,  Anthony B. and Flores,  Mona G. and Zhang,  Ying and Magoc,  Tanja and Harle,  Christopher A. and Lipori,  Gloria and Mitchell,  Duane A. and Hogan,  William R. and Shenkman,  Elizabeth A. and Bian,  Jiang and Wu,  Yonghui},
  year = {2022},
  month = dec 
}

@misc{medgemma,
      title={MedGemma Technical Report}, 
      author={Andrew Sellergren and Sahar Kazemzadeh and Tiam Jaroensri and Atilla Kiraly and Madeleine Traverse and Timo Kohlberger and Shawn Xu and Fayaz Jamil and Cían Hughes and Charles Lau and Justin Chen and Fereshteh Mahvar and Liron Yatziv and Tiffany Chen and Bram Sterling and Stefanie Anna Baby and Susanna Maria Baby and Jeremy Lai and Samuel Schmidgall and Lu Yang and Kejia Chen and Per Bjornsson and Shashir Reddy and Ryan Brush and Kenneth Philbrick and Mercy Asiedu and Ines Mezerreg and Howard Hu and Howard Yang and Richa Tiwari and Sunny Jansen and Preeti Singh and Yun Liu and Shekoofeh Azizi and Aishwarya Kamath and Johan Ferret and Shreya Pathak and Nino Vieillard and Ramona Merhej and Sarah Perrin and Tatiana Matejovicova and Alexandre Ramé and Morgane Riviere and Louis Rouillard and Thomas Mesnard and Geoffrey Cideron and Jean-bastien Grill and Sabela Ramos and Edouard Yvinec and Michelle Casbon and Elena Buchatskaya and Jean-Baptiste Alayrac and Dmitry Lepikhin and Vlad Feinberg and Sebastian Borgeaud and Alek Andreev and Cassidy Hardin and Robert Dadashi and Léonard Hussenot and Armand Joulin and Olivier Bachem and Yossi Matias and Katherine Chou and Avinatan Hassidim and Kavi Goel and Clement Farabet and Joelle Barral and Tris Warkentin and Jonathon Shlens and David Fleet and Victor Cotruta and Omar Sanseviero and Gus Martins and Phoebe Kirk and Anand Rao and Shravya Shetty and David F. Steiner and Can Kirmizibayrak and Rory Pilgrim and Daniel Golden and Lin Yang},
      year={2025},
      eprint={2507.05201},
      archivePrefix={arXiv},
      primaryClass={cs.AI},
      url={https://arxiv.org/abs/2507.05201}, 
}

@misc{gemma,
      title={Gemma: Open Models Based on Gemini Research and Technology}, 
      author={Gemma Team},
      year={2024},
      eprint={2403.08295},
      archivePrefix={arXiv},
      primaryClass={cs.CL},
      url={https://arxiv.org/abs/2403.08295}, 
}

@article{PSNincancer,
  title = {Uncovering the Understanding of the Concept of Patient Similarity in Cancer Research and Treatment: Scoping Review},
  volume = {27},
  ISSN = {1438-8871},
  url = {http://dx.doi.org/10.2196/71906},
  DOI = {10.2196/71906},
  journal = {Journal of Medical Internet Research},
  publisher = {JMIR Publications Inc.},
  author = {Manuilova,  Iryna and Bossenz,  Jan and Weise,  Annemarie Bianka and Boehm,  Dominik and D\"{o}bel,  Marvin and Werle,  Silke D and Ustjanzew,  Arsenij and Reimer,  Niklas and Strantz,  Cosima and Unberath,  Philipp and Metzger,  Patrick and Pauli,  Thomas and Schulze,  Susann and Hiemer,  Sonja and Oguzt\"{u}rk,  Irmak and Kamkar,  Leila and Kestler,  Hans A and Busch,  Hauke and Brors,  Benedikt and Christoph,  Jan},
  year = {2025},
  month = aug,
  pages = {e71906}
}

@article{GNNEHR2024,
  title = {Graph neural networks for clinical risk prediction based on electronic health records: A survey},
  volume = {151},
  ISSN = {1532-0464},
  url = {http://dx.doi.org/10.1016/j.jbi.2024.104616},
  DOI = {10.1016/j.jbi.2024.104616},
  journal = {Journal of Biomedical Informatics},
  publisher = {Elsevier BV},
  author = {Oss Boll,  Heloísa and Amirahmadi,  Ali and Ghazani,  Mirfarid Musavian and Morais,  Wagner Ourique de and Freitas,  Edison Pignaton de and Soliman,  Amira and Etminani,  Farzaneh and Byttner,  Stefan and Recamonde-Mendoza,  Mariana},
  year = {2024},
  month = mar,
  pages = {104616}
}

@article{PSNheart2024,
  title = {A Patient Similarity Network (CHDmap) to Predict Outcomes After Congenital Heart Surgery: Development and Validation Study},
  volume = {12},
  ISSN = {2291-9694},
  url = {http://dx.doi.org/10.2196/49138},
  DOI = {10.2196/49138},
  journal = {JMIR Medical Informatics},
  publisher = {JMIR Publications Inc.},
  author = {Li,  Haomin and Zhou,  Mengying and Sun,  Yuhan and Yang,  Jian and Zeng,  Xian and Qiu,  Yunxiang and Xia,  Yuanyuan and Zheng,  Zhijie and Yu,  Jin and Feng,  Yuqing and Shi,  Zhuo and Huang,  Ting and Tan,  Linhua and Lin,  Ru and Li,  Jianhua and Fan,  Xiangming and Ye,  Jingjing and Duan,  Huilong and Shi,  Shanshan and Shu,  Qiang},
  year = {2024},
  month = jan,
  pages = {e49138–e49138}
}

@misc{captum,
      title={Captum: A unified and generic model interpretability library for PyTorch}, 
      author={Narine Kokhlikyan and Vivek Miglani and Miguel Martin and Edward Wang and Bilal Alsallakh and Jonathan Reynolds and Alexander Melnikov and Natalia Kliushkina and Carlos Araya and Siqi Yan and Orion Reblitz-Richardson},
      year={2020},
      eprint={2009.07896},
      archivePrefix={arXiv},
      primaryClass={cs.LG},
      url={https://arxiv.org/abs/2009.07896}, 
}

\appendix

\section{Ethics statement}
This study was approved by the Institute Ethics Committee. Data were sourced from electronic health records obtained during routine clinical care, and all records were deidentified prior to use to ensure patient privacy. A team of human annotators (two trained nurses and one doctor) was recruited and employed as part of a research project for one year to perform data cleaning, standardization, and deidentification. Ethical working conditions were ensured and all annotators were fairly compensated, receiving salaries (\$0.33 for annotating each note by a nurse and \$1 for annotating each note by the doctor) in accordance with the national guidelines for staff remuneration.
\section{Limitations}
\label{sec:limitations}
While this work advances automated locus of care triaging, it also has a few limitations. \textbf{\textit{First}}, our data set is obtained from a single hospital source, which may hinder the generalizability of \ours{} over other domains. \textbf{\textit{Second}}, the data are limited to clinical notes only for prediction. Future works may include multimodal features such as image (MRI scan to study brain damage by chronic substance use) and audio (to capture the nuances of speech) to model any clinical decision support system. \textbf{\textit{Third}}, \ours{} uses heuristics in the genetic algorithm, which may have caused sub-optimal meta-graph construction. Better approaches may be employed to obtain an optimal, unbiased meta-graph to drive meta learning. \textbf{\textit{Finally}}, while \ours{} aims to provide accurate prediction, it also risks providing incorrect or misleading information at times, and therefore, this framework should always be used as an assistive tool with clinicians-in-the-loop.
\section{Details of annotation}
\label{annotation_instructions}
Each annotator is presented with a single clinical note at a time using \textsc{Label Studio}\footnote{\url{https://labelstud.io/}} for standardization, NER recognition, and target leak masking of the note. Please note that this tool is used in accordance with its intended use. This was conducted by practicing doctors and nurses who were trained by experts in the domain for two weeks. \\
\noindent\textbf{Instruction for standardization of notes}: \\
The instructions to standardize the clinical note are as follows:
\begin{compactenum}
    \item Expand all abbreviations, place the contraction in round brackets after the expansion.
    \item Correct spelling mistakes, punctuation errors, and capitalization errors (for example, abbreviations may not be capitalized – ``seen in opd'' to ``Seen in Out Patient Department (OPD)''.
    \item Break down long sentences into multiple sentences, even if it becomes agrammatical. Make sure to preserve meaning of the original sentence.
    \item End each sentence with a period and capitalise the first letter of each sentence.
    \item Arrange sentences into paragraphs based on similar themes of information. Then arrange the paragraphs into sections as follows:
    \begin{compactenum}
        \item Paragraph - all sentences containing information about current alcohol use.
        \item Paragraph - all sentences containing information about current tobacco use.
        \item Paragraph - all sentences containing information about illnesses in the biological relatives.
        \item Sections - The paragraphs containing information about current alcohol and tobacco use are then placed in the section ``History of presenting illness''; the paragraph on illnesses of family members is placed in the section ``Family history''.
    \end{compactenum}
    \item Retain numerals, dates, medication dosages without change.
\end{compactenum} 

\noindent\textbf{Instructions for NER recognition in notes}: \\
The note will contain various words which can lead to the identification of the patient. We need to select and correctly categorize this information into entities. The list of named entity types and their descriptions are given below.
\begin{compactenum}
    \item Person: Name of persons with their prefixed salutations: Eg: Dr LS, Alice, Dan, etc.
    \item Company: Name of companies or organizations. This includes names of healthcare facilities or any institutions which can be employers. Extract only if there is a name to the entity which can lead to identification: for example ``government hospital'' is a generic term and need not be included.
    \item Language: Name of languages.
    \item Dates: Include only fully specified dates which include Day, Month and Year.
    \item Address: Names of countries (address country), names of states (address states) and all other locations or geographical entities which are smaller than a state (address).
    \item Identification number: Numeric or alphanumeric identifiers which can be directly linked to an individual, including phone numbers, driving license, hospital identity number etc.
    \item Groups: Names of groups which can bias or lead to identification, including religion, caste, tribes, political groups and self-help groups. Eg: ``Muslim'', ``hakki pikki'', etc.
\end{compactenum}
Select the words or phrases corresponding to each entity, then click on the entity type you want to assign it to. The selected word or phrase will then get highlighted. Like this, highlight and assign all identifying words and phrases in the document. Once you reach the end of the document, review the note to ensure that all identifying information has been covered, then submit.

\noindent\textbf{Instruction for target leak masking of notes}:\\
Follow the instruction guidelines given to annotate the masking of target leakage in a note.
\begin{compactenum}
    \item All phrases or sentences which indicate treatment or management decisions must be selected. This includes what medications were prescribed, whether the patient was admitted to ward, or referred to another specialist or hospital. Eg: ``Send to ward'', ``Admit'', ``To be seen in a week'', ``May benefit with admission'', ``Needs observation'', ``Detox to be done'', ``Regular compliance and follow up'', etc.
    \item We only need information that reveals what happened at the end of this consultation, not the past.
    \item We only need information related to admission and medications, not general advice or psychological interventions.
    \item We only need information which reveals doctors plans and not patient or families' requests. For example, ``Beds not available at present'' or ``Declined IP care'' reveals that the doctor wanted to admit whereas ``Patient is requesting admission'' does not reveal information about the doctor’s decision and need not be selected.
\end{compactenum}
\section{Further analysis of the dataset}
\label{sec:dataset_analysis}
\begin{figure}[t]
  \includegraphics[width=\linewidth]{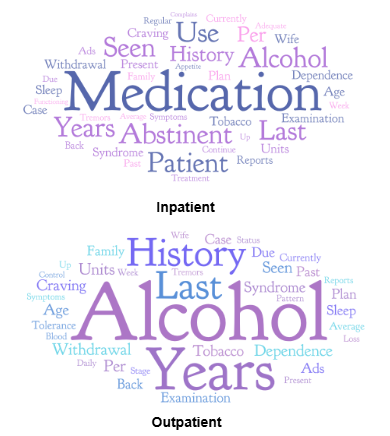}
  \caption{Word clouds of frequently occurring words in IP and OP classes.}
  \label{fig:wordcloud}
\end{figure}

In this section, we report a few more statistics of the dataset used in this study. The average length of $\mathcal{N}$ is approximately 244 words, with a maximum going up to 580. Figure~\ref{fig:wordcloud} represents the word clouds of frequently co-occurring words for each class present in $\mathcal{N}$. As we can see, words like \textit{alcohol}, \textit{dependence}, \textit{withdrawal}, etc., are common in both classes, signifying the challenging task of binary classification. Therefore, we used trigrams as an indicative feature to separate between classes.\\
\noindent\textbf{PII removal performance}: Out of the 8513 entities obtained from 1850 annotated $\mathcal{N}$, the most common entities are \textit{address}, with 2736 instances, and \textit{person} with 2637. Other significant entities include \textit{company} and \textit{dates} with 2072 and 816 occurrences, respectively. Finally, a smaller number of entities were classified as \textit{groups}, \textit{languages} and \textit{numerical identifiers} with 124, 110 and 14 occurrences, respectively.
The character-level performance of entity detection on 350 held-out annotated entries from $\mathcal{N}$ is reported in Table~\ref{tab:entity_perf}.
\begin{table}[htbp]
\centering
\begin{tabular}{lccc}
\toprule
\textbf{PII entity} & \textbf{Precision} &\textbf{Recall} & $\mathbf{F_1}$ \\
\midrule
Person & 0.99 & 0.97 & 0.98 \\
Languages & 1.00 & 0.91 & 0.95 \\
Groups & 1.00 & 1.00 & 1.00 \\
Company & 0.94 & 0.97 & 0.96 \\
Dates & 0.98 & 0.99 & 0.98 \\
Numerical ID & 0.88 & 0.82 & 0.85 \\
Address & 1.00 & 1.00 & 1.00 \\
\bottomrule
\end{tabular}
\caption{Character level performance for PII entities.}
\label{tab:entity_perf}
\end{table}

\noindent\textbf{PSN statistics}:
We obtained a separate train and test patient similarity network. The training graph consists of 4,988 nodes and 120,492 edges that connect similar patients. The test graph contains 2,640 nodes and 69,600 edges. The average number of edges per node is approximately 45 in both graphs. The average degree of nodes in IP and OP classes is 39 and 52, respectively. The number of isolated nodes is 1353 and 622 in the train and test graphs, respectively.
\section{Implementation details}
Our implementation was developed using Python 3.10 with PyTorch 2.0 and PyTorch Geometric \textbf{(PyG)} as the primary deep learning framework. All models were trained on an NVIDIA A6000 GPU with four core and 48GB memory, utilizing CUDA 12.2 for accelerated computation.

\subsection{Genetic algorithm parameter optimization}
We used genetic algorithm (GA) to construct an unbiased meta-graph for meta learning. The parameters of the genetic algorithm, including population size, number of generations, crossover rate, and mutation rate are tuned using Optuna \cite{optuna_2019} and the meta-graph that has the best fitness score is selected. The GA parameters are optimised and set as population size = 50, generations = 100, crossover rate = 0.8745703676281257, and mutation rate = 0.21873583752075254. The population size and number of generations are explicitly chosen to select low values for quick and efficient computation. Increasing the population size and number of generations beyond 300 was computationally inefficient.
\subsection{Fine-tuning Configuration}
\label{app:llmft}
\subsubsection{\textsc{GatorTron}} We fine-tuned \textsc{GatorTron} end-to-end foloowed by a classification head at the final layer for binary classification task. The fine-tuning pipeline integrates the pre-trained \textsc{GatorTron} transformer encoder to extract contextual feature extractor with two-layer feedforward network to project latent embeddings. Since the input context length is limited to 2048, we chunked and aggregated longer notes with mean pooling. The training is done by the utilization of AdamW with different learning rates for the base and head layers and a OneCycleLR scheduler for the adaptation of the rate. The loss function, BCEWithLogitsLoss, includes class imbalance correction through the weighting of the positive class. In addition, we use gradient accumulation, early stopping for further optimization. The fine-tuning is with EARLY\_STOPPING\_PATIENCE = 2, BASE\_LR (learning rate for \textsc{GatorTron} base model) = 2e-5, HEAD\_LR (learning rate for classification head) = 1e-4 , WEIGHT\_DECAY = 0.01, and WARMUP\_RATIO = 0.1 over 10 epochs.
\subsubsection{\textsc{MedGemma}}
Due to hardware constraints and unavailability of hosted services for this model, we had to perform inference in 4-bit NF4 quantisation. The quantisation was performed with standard BitsAndBytesConfig arguments. A LoRA adapter of rank 16, scaling factor 16, dropout of 0.05, and target\_modules=``all-linear'' is utilized to all linear layers, limiting the number of trainable parameters to low-rank. The custom DataCollator to implement the assistant-only loss by utilizing the padding of input\_ids, the creation of attention masks, and the duplicating of input\_ids to labels. The loss is computed only on the IP/OP answer tokens excluding the user prompt. The model was required to output `IP' or `OP'. If it gave a response which did not have these keywords or had both, we retried for 5 times before counting that case as a failure.
\section{Extended ablation study}
\label{apx:ablation} 
\begin{figure}[!htpb]
    \centering
    \includegraphics[width=\linewidth]{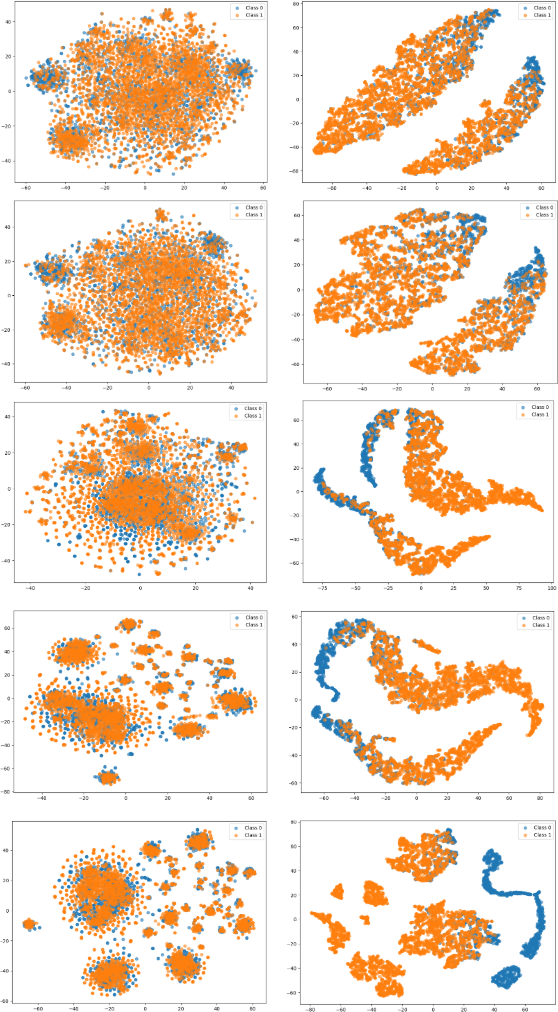}
    \caption{Side-by-side comparison of $t$-SNE plots of node embeddings before and after training with \ours{}. The rows represents embeddings used as \fb{}, \fb{}+\fl{}, \fb{}+\fe{}, \fb{}+\fe{}+\fl{} and \fb{}+\fe{}+\fl{}+\fr{} respectively. Here blue and orange dots indicate IP and OP classes respectively. }
    \label{fig:tsne_ipop}
\end{figure}

We further extend the ablation study of \ours{} in this section. To support the results in Table~\ref{tab:encoding_ablation}, we draw $t$-SNE plots for each of the components of the node embedding in Figure~\ref{fig:tsne_ipop}.

\section{Error analysis}
\label{apx:error_analysis}

\begin{table*}[!h]
\footnotesize
    \centering
    \begin{tabular}{p{0.40\linewidth}|p{0.05\linewidth}|p{0.05\linewidth}|p{0.30\linewidth}}
    \toprule
     \textbf{Clinical note} & \textbf{\ours{}} & \textbf{Ground truth} & \textbf{Observation} \\
    \midrule
    Patient seen with aunt, brother \{person\_01\}, 50-year-old male, Pollution control board, single, \{address\_01\}...Patient usually binges for 1 week or month and sometimes more...  Patient has \sethlcolor{lightgreen}\hl{severe vomiting following these binges. Possibly Mallory-Weiss tear-related frank blood in vomitus (no coffee ground/non-bilious).} Patient has occasional simple withdrawal symptoms. Patient drinks about 750 mL on average per sitting. Has been functional throughout. Last week had an \sethlcolor{lightgreen}\hl{episode of suspected seizure for 2 minutes.} ...  Treatment\_decision3 start Treatment\_decision2 patient does not have any withdrawal Treatment\_decision1 currently.  However, \sethlcolor{yellow}\hl{patient has problems with anger management and mood.} Therefore, to \sethlcolor{yellow}\hl{address that in the next follow-up.} ... Complains of (C/O): \sethlcolor{yellow}\hl{no fresh complaints. Supportive work done.} Discussed With (D/W) \{person\_01\}, Senior Resident (SR), \{company\_01\}. \sethlcolor{yellow}\hl{Abstinent from alcohol.} Current concern is sleep disturbance. Difficulty initiating and maintaining sleep. & OP & IP & \ours{} ignores the fact that the patient had severe vomiting with blood and had a suspected seizure. These conditions require intensive evaluation and close monitoring, necessitating inpatient care. It focuses on less consequential and low-emergency points such as anger management and mood. \\
    \midrule
    \{person\_02\}, 39 years old, Male, Married, 8th standard pass, Driver, Resident of \{address\_02\}... Increased in frequency and quantity since 1.5 years. \sethlcolor{yellow}\hl{Associated with craving, tolerance, loss of control, and withdrawal symptoms in the form of tremulousness and sleep disturbances. With average use of 6 to 18 units per day. With last use around 18 units at 2 AM yesterday.} Relapse due to craving and secondary to interpersonal relationship issues with wife. Wife lives separately from patient since 2 years along with their children... \sethlcolor{lightgreen}\hl{No history of psychotic symptoms.} It seems that patient gravitates towards alcohol to seek relief when he is undergoing stressful times in his life. Maintaining factors seem to be craving, loss of control and interpersonal relationship issues with wife.  On Examination:  \sethlcolor{lightgreen}\hl{Conscious, Oriented.  Pulse Rate (PR) is 105 per minute.  Blood Pressure (BP) is 135/88 mm Hg.  Body Mass Index (BMI) is 21.58 kg/m². Motivation is in Preparatory stage for alcohol cessation and Contemplation stage for tobacco cessation.  Mental Status Examination (MSE):  Euthymic affect.}  Provisional Impression:  Alcohol Dependence Syndrome (ADS) with Simple Withdrawal State (SWS). Management Plan: 1)Treatment\_decision4 and gastro protective measures to be ensured. 2)Treatment\_decision3 can be initiated as it seems that patient is a relief drinker. 3)Treatment\_decision2 can be initiated.Treatment\_decision1. 4) Relapse Prevention Therapy (RPT) can be initiated. 5) To check if Psychiatric Social Worker (PSW) team can contact wife as current worsening due to apparent interpersonal relationship issues with wife. 6) To follow up after 2 weeks. Case Seen By (C/S/B) \{person\_01\}, Consultant under \{company\_01\}. Plan:  1) Detoxification to be initiated.  2) \sethlcolor{lightgreen}\hl{To follow up after 1 week and to plan for further management after detoxification.} & IP & OP & \ours{} focuses on the presence of withdrawal symptoms and average use of alcohol. However, on physical examination, the patient was in simple withdrawal and was highly motivated to stop alcohol, which warrants outpatient care.\\
    \bottomrule
    \end{tabular}
    \caption{A couple of examples of clinical note misclassified by \ours{}. The text \sethlcolor{yellow}\hl{highlighted in yellow} represents the words which was focused by \ours{}, while text \sethlcolor{lightgreen}\hl{highlighted in green} indicates the potential words/phrases which \ours{} should also have attended.} 
    \label{tab:grace_misclassification}
\end{table*}

Beyond evaluation metrics, we analyze misclassification patterns to make the model more reliable. 
We present a couple of examples of clinical patient notes in Table~\ref{tab:grace_misclassification} where \ours{} fails to capture the underlying semantics and context of medical naunces.

\section{Prompts}
\label{apx:prompts}
\begin{figure*}[t]
  \includegraphics[width=\linewidth]{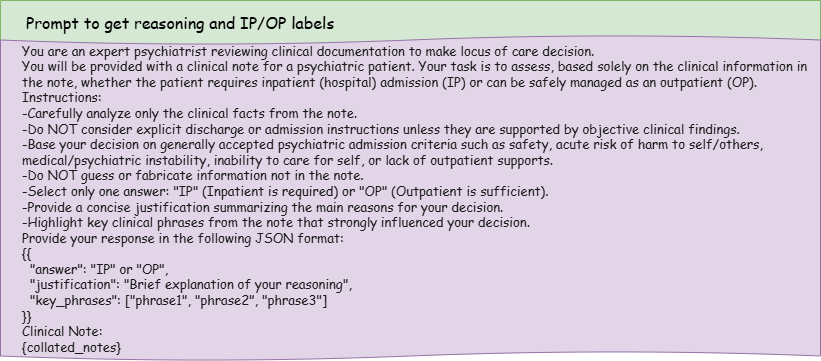}
  \caption {Prompt used to get classification prediction along with reasoning pathways.}
  \label{fig:prompts}
\end{figure*}

The prompts used in zero-shot setting to obtain the final prediction with their reasoning pathways for each of the LLM based baselines are described in Figure~\ref{fig:prompts}. 

\end{document}